\pdfoutput=1
\documentclass[11pt]{article}

\usepackage[]{acl}

\usepackage{times}
\usepackage{latexsym}

\usepackage[T1]{fontenc}
\usepackage[utf8]{inputenc}

\usepackage{microtype}

\usepackage{multirow, makecell}
\usepackage{booktabs}
\usepackage{amsmath}
\usepackage{graphicx}

\title{Improving Top-K Decoding for Non-Autoregressive Semantic Parsing via Intent Conditioning}

\author{
    Geunseob (GS) Oh, Rahul Goel, Chris Hidey, \\ 
    \textbf{Shachi Paul, Aditya Gupta, Pararth Shah, Rushin Shah} \\
    Google \\
    \texttt{\{\small ohgs,goelrahul,chrishidey,shachipaul,gaditya,pararth,rushinshah\}@google.com}
}

\begin{document}
\maketitle
\begin{abstract}

Semantic parsing (SP) is a core component of modern virtual assistants like Google Assistant and Amazon Alexa. While sequence-to-sequence based auto-regressive (AR) approaches are common for conversational SP, recent studies \citep{ref:fb_span_pointer_network} employ non-autoregressive (NAR) decoders and reduce inference latency while maintaining competitive parsing quality. 
However, a major drawback of NAR decoders is the difficulty of generating top-\emph{k} (i.e., \emph{k}-best) outputs with approaches such as beam search. 
To address this challenge, we propose a novel NAR semantic parser that introduces intent conditioning on the decoder. Inspired by the traditional intent and slot tagging parsers, we decouple the top-level intent prediction from the rest of a parse. 
As the top-level intent largely governs the syntax and semantics of a parse, the intent conditioning allows the model to better control beam search and improves the quality and diversity of top-\emph{k} outputs. We introduce a hybrid teacher-forcing approach to avoid training and inference mismatch. We evaluate the proposed NAR on conversational SP datasets, TOP \& TOPv2. Like the existing NAR models, we maintain the $O(1)$ decoding time complexity while generating more diverse outputs and improving top-3 exact match (EM) by $2.4$ points. In comparison with AR models, our model speeds up beam search inference by $6.7$ times on CPU with competitive top-\emph{k} EM.  

\end{abstract}

%%%%%%%%%%%%%%%%%%%%%%%%%%%%%%%%%%%%%%%%%%%%%%
\section{Introduction}
\label{sec:Introduction}
%%%%%%%%%%%%%%%%%%%%%%%%%%%%%%%%%%%%%%%%%%%%%%

% 1. Background: recent seq2seq semantic parsing

\noindent Neural sequence models are widely used for the task of conversational semantic parsing, which converts natural language utterances to machine-understandable meaning representations. Recent approaches \citep{ref:TOPv2, ref:amazon_dont_parse_generate,  ref:NSP_retrain, ref:SBTOP, ref:fb_span_pointer_network} combine Transformer-based sequence models \citep{ref:Transformer, ref:BERT} and Pointer Generator Networks \citep{ref:pointer_network, ref:pointer_generator_network}. A vast majority of semantic parsing approaches \citep{ref:TOP, ref:TOPv2, ref:amazon_dont_parse_generate, ref:NSP_retrain, ref:SBTOP, ref:compositional_span_level_attention} employ autoregressive (AR) decoders to generate structured output frames for the quality of output parses. During the AR decoding, output tokens are generated sequentially, conditioned on all previously generated tokens. As a result, the decoding time (i.e. inference latency) increases linearly with the decoding length. This not only limits the capacity to accommodate larger language models but may also degrade the user experience of intelligent conversational assistants (e.g. Google Assistant, Amazon Alexa) due to the high inference latency.

\begin{table}[t]
\begin{tabular}{
>{\arraybackslash}m{0.7cm}
>{\arraybackslash}m{6.15cm} }
\toprule
\multicolumn{2}{l}{Query: what is going on this weekend?} \\
Label: & [in:get\_event [sl:date\_time this weekend ] ] \\
\midrule
\multicolumn{2}{l}{(Failure 1: repeated tokens)} \\
& [in:get\_event [sl:date\_time this weekend ] [sl:date\_time this weekend ] ] \\
& (Parse with the repeated slot \textcolor{purple}{\textit{sl:data\_time}} and leaf nodes \textcolor{purple}{\textit{this}} and \textcolor{purple}{\textit{weekend}}) \\
\multicolumn{2}{l}{(Failure 2: invalid syntax)} \\
& [in:get\_event [sl:date\_time \\
& (Incomplete parse; missing `\textcolor{purple}{] ]}') \\
\bottomrule
\end{tabular}
\caption{Examples of possible failures from the limited beam search of the existing NAR semantic parsers.}
\label{Table:baseline_NAR_failure}
\end{table}

% 2. Non-autoregressive semantic parsing is useful; low latency, on-device deployment etc
% 3. Recent NAR models have shown promising results (cite the Span Pointer Network)

On the contrary, non-autoregressive (NAR) decoders are capable of parallel decoding and thus allow much faster inference. Compared to AR decoders, which perform O($n$) decoding steps, the NAR decoding is typically achieved in either $O(1)$ \citep{ref:fb_NAR_lightconv_pointer, ref:fb_span_pointer_network} or O($\mathrm{log}(n)$) \citep{ref:amazon_dont_parse_insert} steps. A recent study \citep{ref:fb_NAR_lightconv_pointer} built NAR parsers and reduced the latency up to 81\% compared to AR parsers. Another study by \citet{ref:fb_span_pointer_network} proposed a NAR semantic parser called Span Pointer Network and cut the latency by 8.5-10x on CPU while achieving comparable performance to their AR benchmarks. All these studies suggest that the NAR decoders have significant latency benefits.

% 4. Limitation of the existing NAR models: the beam search \& topk outputs are limited (lack of diversity, poor % performance). This is because the current NAR models are based on frame-length conditioning. 

Although the recent NAR semantic parsers have decreased the performance gap from AR parsers on the exact match metric, one of the main limitations of NAR models still remain unsolved: beam search and top-\emph{k} outputs. AR models leverage beam search algorithms \citep{ref:beam_is_effective_for_seq2seq}, which are especially effective at producing top-\emph{k} outputs \citep{ref:beam_is_effective_for_seq2seq, ref:TOP}. This is possible since the beam search allows AR models to dynamically sort out less-probable candidate parses and adjust their decoding lengths. As a result, AR models are capable of generating diverse high-quality output parses. 

% \Shachi{Shachi: we may want to draw a parallel between how the parses generated by AR parsers are more likely to be valid than NAR parsers courtesy the conditioning. We do mention the "invalid" parses bit for NAR but have missed the parallel comparison with AR above}
% \GS{GS: edited the following paragraphs for Shachi's suggestion, but please feel free modify or comment!} 
On the other hand, the existing NAR parsers \citep{ref:fb_NAR_lightconv_pointer, ref:fb_span_pointer_network} cannot employ the AR beam search as the output tokens of the NAR parsers are generated independently from each other. Instead, the existing NAR parsers perform a beam search by generating $k$ candidate frame lengths. For each frame length, a single parse is produced, resulting in a total of $k$ parses.

% \ag{can we itemize the issue with naive beam search}
However, we find that the existing NAR beam search tends to produce duplicates of the most probable output parse. As exemplified in Table \ref{Table:baseline_NAR_failure}, top-\emph{k} beam outputs often include outputs with repeated tokens and parses with invalid syntax (e.g., truncated or extended versions of the most-probable parse) rather than diverse output parses. 

The ability to generate \emph{diverse} top-\emph{k} parses is important for modern conversational assistants. Table \ref{Table:proposed_NAR_example_output} presents an example that demonstrates how a query may correspond to different parses depending on the context. The diverse semantic parses can be leveraged in a downstream component that has more contextual information. An additional re-ranking module can also be employed to select the most relevant semantic parse.

\begin{table}[t]
\begin{tabular}{
>{\arraybackslash}m{0.35cm}
>{\arraybackslash}m{6.5cm} }
\toprule
\multicolumn{2}{c}{Query: avoid bridges on my route} \\
\midrule
\multicolumn{2}{c}{Parses from the Proposed NAR} \\
(1a) & \textcolor{cyan}{[in:update\_directions} [sl:path\_avoid [in:get \\
& \_location [sl:category\_location bridges\textcolor{brown}{]]]]} \\
(2a) & \textcolor{blue}{[in:get\_directions} [sl:path\_avoid [in:get \\
& \_location [sl:category\_location bridges\textcolor{brown}{]]]]}\\
\midrule
\multicolumn{2}{c}{Parses from the Baseline NAR} \\
(1b) & \textcolor{cyan}{[in:update\_directions} [sl:path\_avoid [in:get \\
& \_location [sl:category\_location bridges\textcolor{brown}{]]]]} \\
(\textcolor{red}{2b}) & \textcolor{cyan}{[in:update\_directions} [sl:path\_avoid [in:get \\
& \_location [sl:category\_location bridges \\
\bottomrule
\end{tabular}
\caption{Examples of beam outputs. The parses produced by the proposed NAR are both valid depending on the context. If the query was made after the user already had set a path, parse 1a is more relevant, else 2a. In contrast, the baseline NAR only produced one valid parse. Parse 2b is an invalid duplicate of parse 1b.}
\label{Table:proposed_NAR_example_output}
\end{table}

% [in:update_directions [sl:path_avoid [in:get_location [sl:category_location bridges bridges ] ] ] ]
% [in:update_directions [sl:path_avoid [in:get_location [sl:category_location bridges bridges

% 5. Summary of our contributions: intent-conditioning improves the performance of NAR models for greedy decoding, especially for beam decoding. Mention that we conducted experiments on TOP, TOPv2, MTOP.

This work focuses on improving the quality of top-\emph{k} outputs of NAR conversational semantic parsing. Our idea is to leverage the fact that the top-level intent of a semantic parse mainly determines the syntax and semantics of the parse. This is the key driving point of the classic intent classification and slot tagging \citep{ref:intent_slot_tagging} based parsers, which were widely used traditional semantic parsers. We build our model upon the existing NAR (i.e., frame length conditioned NAR), but decouple the prediction of the top-level intent from the output. We leverage the conditional dependencies of the frame length and parse on the top-level intent for diverse top-\emph{k} outputs. 

We evaluate the proposed NAR model on two datasets: TOP \citep{ref:TOP} and TOPv2 \citep{ref:TOPv2} and demonstrate that we further close the gap between AR and NAR parsers. Most importantly, we show that the intent conditioning improves the quality and diversity of the top-\emph{k} parses and the total inference time of our NAR beam search is 6.7x times shorter than the AR baseline on CPU.

% Below maybe used somewhere

% We discovered that only small percentage of the mismatched predictions had incorrect frame length. Many of the mismatched predictions instead had incorrect top-level intent predictions. % TODO: provide the numbers

%%%%%%%%%%%%%%%%%%%%%%%%%%%%%%%%%%%%%%%%%%%%%%
\section{Related Work}
\label{sec:related_works}
%%%%%%%%%%%%%%%%%%%%%%%%%%%%%%%%%%%%%%%%%%%%%%

\subsection{Non-autoregressive Semantic Parsing}

% 1. NAR NSP models: Transformers + Pointer Generator Network. The existing models are frame length conditioned ones.

NAR sequence models have been an active research area across different fields of natural language processing for their fast inference speeds. While various designs of NAR decoders exist, recent works in machine translation use models with iterative refinements \citep{ref:MT_NAR_iterative_refinement, ref:MT_NAR_mask_predict, ref:MT_NAR_mask_predict_semi_autoregressive}, insertion-based \citep{ref:MT_NAR_insertion}, and latent alignment methods \citep{ref:MT_NAR_CTC, ref:MT_NAR_Imputer, ref:MT_NAR_latent_alignment}.

The task of semantic parsing (SP) is similar to the machine translation task as they both translate input sentences from one representation to another. For this reason, recent studies in SP have adopted modeling techniques from machine translation. \citet{ref:amazon_dont_parse_insert} leveraged the insertion-based seq2seq models for NAR semantic parsing and reduced the decoding time from $O(n)$ to $O(log(n))$ while matching the performance of AR models. \citet{ref:fb_NAR_lightconv_pointer, ref:fb_span_pointer_network} applied an iterative refinement method \emph{Mask-Predict} to semantic parsing and brought the time complexity further down to $O(1)$ while performing comparably to the baseline AR models. As opposed to the original Mask-Predict model that performs multiple iterations of re-masking and prediction \citep{ref:MT_NAR_mask_predict}, they only perform a single iteration of the masking and output prediction as they claim that task-oriented SP does not benefit much from iterative refinements \citep{ref:fb_NAR_lightconv_pointer, ref:fb_span_pointer_network}. In addition to the recent sequence-to-sequence NAR semantic parsers, traditional intent and slot filling model \citep{ref:rnn_slot_filling} is another approach with non-autoregressive decoding.

\subsection{Baseline NAR}

Span Pointer Network \citep{ref:fb_span_pointer_network}, which is one of the two recent works that leveraged the mask-predict algorithm, showed the competitive performance to the AR parsers while significantly reducing their latency. For this reason, we utilize Span Pointer Network as a basis of our work and accordingly refer to it as \emph{baseline NAR} to distinguish the NAR semantic parser we propose. It should be noted that we replicated Span Pointer Network as it is not publicly available. 

The baseline NAR is built with a pre-trained encoder, a frame length module, and a Transformer decoder. The encoder is a language model such as BERT \citep{ref:BERT} or RoBERTa \citep{ref:RoBERTa} that takes input queries $x$ and outputs their encoded representations $e$. 

\begin{equation}
\label{eq:baseline_nar_encoder}
\begin{aligned}
e_{1:l} = \mathrm{Encoder}(x_{1:l}),
\end{aligned}
\end{equation}

where $l$ indicates the length of the source (i.e., input query). The frame length module takes the encoded representations and predicts the length of the frame (i.e., target parse) $n$ followed by the generation of $n$ mask tokens.

\begin{equation}
\label{eq:baseline_nar_length_module}
\begin{aligned}
n = \mathrm{FrameLengthModule}(e_{1:l}). \\
\mathrm{[MASK]}_{1:n} = \mathrm{MaskCreation}(n).
\end{aligned}
\end{equation}

The transformer decoder takes the $N$ mask tokens as inputs and produce $h$ for each token.

\vspace{-15pt}
\begin{equation}
\label{eq:baseline_nar_decoder}
\begin{aligned}
h_{1:n} = \mathrm{Decoder}(\mathrm{[MASK]}_{1:n}; e_{1:l}).
\end{aligned}
\end{equation}

Lastly, the pointer-generator mapping layer is used to convert $h$ to the target frame $y$ as follows.

\vspace{-5pt}
\begin{equation}
\label{eq:baseline_nar_ptr}
\begin{aligned}
y_{1:n} = \mathrm{PTR}(h_{1:n}; e_{1:l}),
\end{aligned}
\end{equation}
where $y_{k}$ is either a token from the target vocab that consists of parse symbols (e.g., intents \& slots) or copy of a source token.
The output tokens are generated in parallel as they are conditionally independent given the source and frame length.

\subsection{Beam search for AR vs baseline NAR}

% How is beam search \& topk decoding done currently with frame-length conditioned models:

AR decoders produce multiple candidate output parses per source by employing the beam search algorithm \citep{ref:beam_is_effective_for_seq2seq}. During the beam search with width $k$, the \emph{k} most probable candidates are kept at each decoding step and less probable candidates are sorted out. The decoding process finishes when a special ``[END]'' token is encountered. For a decoding length $n$, the score of an AR parse is computed as follows.

\vspace{-10pt}
\begin{equation}
\label{eq:ar_scoring}
\begin{aligned}
S_{AR}(y) = p(y|x) = \prod^{n}_{j=1} p(y_{j}|y_{1:j-1},x).
\end{aligned}
\end{equation}

Since the AR decoders consider multiple candidate tokens $y_{j}$ at each decoding step, they can generate a large number of unique parses.

% equation

\begin{figure}[t]
    \centering
    \includegraphics[width=0.85\linewidth]{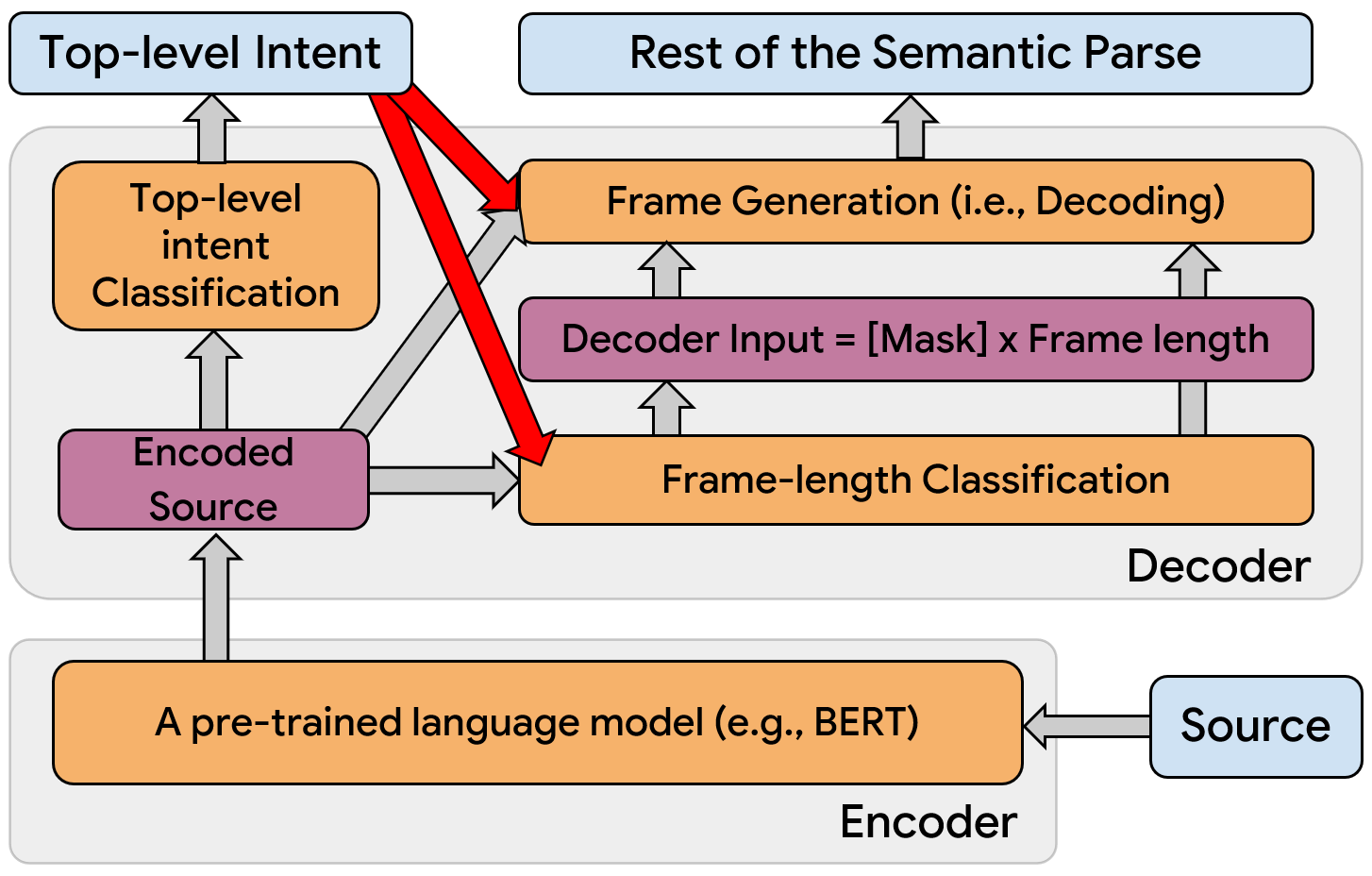}
    \caption{The proposed model architecture. By decoupling the top-level intent prediction from the output prediction, we perform conditional generation of the frame length and parse upon the top-level intent.}
    \label{fig:proposed_model_decouple_intent}
\end{figure}

In comparison, the baseline NAR only produces a single output parse per frame length. The baseline NAR mimics the beam search by generating top-\emph{k} frame lengths, which results in $k$ outputs. The output parse is scored using the joint probability of the frame length and output parse as follows.

\begin{equation}
\label{eq:baseline_nar_scoring}
\begin{aligned}
S_{NAR,baseline}(y) = p(y,n|x) = \\
p(y|n,x) \cdot p(n|x) = \prod^{n}_{j=1} p(y_{j}|n,x) \cdot p(n|x),
\end{aligned}
\end{equation}

where $S_{NAR,baseline}(y)$ denotes the score of the baseline NAR output. $p(y|n,x)$ is obtained using the conditional independence of the output tokens given the frame length $n$ and source $x$.

%%%%%%%%%%%%%%%%%%%%%%%%%%%%%%%%%%%%%%%%%%%%%%
\section{Proposed Approach}
\label{sec:proposed_approach}
%%%%%%%%%%%%%%%%%%%%%%%%%%%%%%%%%%%%%%%%%%%%%%

To mitigate the aforementioned limitation of the baseline NAR, we propose Intent-conditioned Non-autoregressive Neural Semantic Parser. Our approach is motivated by traditional semantic parsers such as \citet{ref:intent_slot_tagging}, which performs an intent classification followed by slot tagging. Another motivation comes from an observation that the top-level intent of the parse largely governs the syntax and semantics of the rest of the parse. 

Our idea is to decouple the top-level intent prediction from the output prediction as shown in Figure \ref{fig:proposed_model_decouple_intent}. This builds explicit dependencies of the frame length and the rest of the frame on the top-level intent. As a result, the joint probability of output and frame length is expressed as follows.

\vspace{-15pt}
\begin{equation}
\label{eq:proposed_nar_joint_prob}
\begin{aligned}
p(y,n|x) = p(y_{2:n}|n,y_{1},x) \cdot p(n|y_{1},x) \cdot p(y_{1}|x).
\end{aligned}
\end{equation}

This facilitates effective conditional generation of the length and output, while keeping the decoding time complexity as $O(1)$; the number of decoding steps does not scale with the output length.

\subsection{Intent Conditioning}
%3. Intent-conditioning model

The proposed NAR utilizes a pre-trained language model to obtain encoded source representation $e$. Then, it performs the top-level intent prediction, which is formulated as a multiclass classification of the size of the dimensionality of intent vocab.

\begin{equation}
\label{eq:proposed_nar_intent_conditioning}
\begin{aligned}
\mathrm{logits}(y_{1}) = \mathrm{IntentModule}(e_{1:l}),
\end{aligned}
\end{equation}

where $y_{1}$ refers to the top-level intent of the semantic parse or the first token of the parse. As the top-level intent prediction is decoupled, the intent module works with a smaller vocabulary size. Compared to the output vocabulary that combines a target and copy index vocabulary, the intent vocabulary is approximately 4 times smaller with a source length 32 on TOP dataset. We use a randomly initialized Transformer as the intent module.

We use the logits of the top-level intent, as opposed to the discrete intent class variable, as the inputs to all subsequent modules. This is because the dense logits convey information about the uncertainty of the intent prediction for each intent class and help the subsequent modules to condition on richer information. We found that this resulted in better performance than the model conditioned on the 1-dimensional discrete intent.

Subsequent to the top-level intent prediction, the frame length module takes the logits of the intent $y_{1}$ and produces the decoding length followed by the creation of mask tokens. After that, the initial mask tokens together with the frame length pass through a positional encoding layer to compute the encoded mask tokens, $\mathrm{[MASK]}_{2:n}$.

\vspace{-10pt}
\begin{equation}
\label{eq:proposed_nar_length_module}
\begin{aligned}
n-1 = \mathrm{FrameLengthModule}(\mathrm{logits}(y_{1}); e_{1:l}), \\
\mathrm{[MASK]}_{2:n} = \mathrm{PosEncoding}(\mathrm{MaskCreation}, n). \\
\end{aligned}
\end{equation}

The rest of the frame is generated using parallel decoding followed by the source token mapping via the pointer-generator mapping layer.

\vspace{-5pt}
\begin{equation}
\label{eq:proposed_nar_decoding}
\begin{aligned}
h_{2:n} = \mathrm{Decoder}(\mathrm{[MASK]}_{2:n}; \mathrm{logits}(y_{1}), e_{1:l}), \\
y_{2:n} = \mathrm{PTR}(h_{2:n}; e_{1:l}), \hspace{5pt}  y_{1:n} = \mathrm{cat}(y_{1}, y_{2:n}).
\end{aligned}
\end{equation}

Finally, the output parse is obtained by concatenating the top-level intent and rest of the frame.

\subsection{Training Objective of the Proposed NAR}
The loss is a weighted sum of top-level intent classification loss $\mathcal{L}_{int}$, frame length classification loss $\mathcal{L}_{len}$, and output loss $\mathcal{L}_{out}$ as follows.

\vspace{-10pt}
\begin{equation}
\label{eq:proposed_nar_loss}
\begin{aligned}
\mathcal{L} = \mathcal{L}_{out} + \lambda_{len} \cdot \mathcal{L}_{len} + \lambda_{int} \cdot \mathcal{L}_{int}.
\end{aligned}
\end{equation}

We jointly optimize the three loss terms and employ label smoothing (LS) \citep{ref:label_smoothing} as a regularizer to penalize overconfident predictions by computing negative log-likelihood (NLL) between the smoothed one-hot labels and predictions. That is, $\mathcal{L}_{int} = \mathrm{NLL}(\mathrm{LS}(y_{1,label}), y_{1,pred})$,  $\mathcal{L}_{len} = \mathrm{NLL}(\mathrm{LS}(n_{label}), n_{pred})$, and $\mathcal{L}_{out} = \mathrm{NLL}(\mathrm{LS}(y_{2:n,label}), y_{2:n,pred})$.

\subsection{Beam Search with the Proposed NAR}
%4. beam search with the intent-conditioned models

The proposed intent-conditioned NAR produces multiple output parses per source via the sequential conditioning on the top-level intent and length. Precisely, the model first computes top-$k1$ top-level intents. For each top-level intent, the length model outputs top-$k2$ frame lengths. As a result, $k1 \cdot k2$ pairs of top-level intents and lengths are produced. Finally, the decoder generates a single parse per pair, yielding a total of $k1 \cdot k2$ output parses.

The proposed NAR performs a single decoder pass regardless of the beam size, $k1 \cdot k2$. This is possible as we cast the $k1 \cdot k2$ pairs as the batch dimension. As a result, the time complexity of the beam search remains constant, to the extent of the memory capacity. The memory increases linearly to the beam size. 

The intent conditioning provides explicit control of the top-level intents via the selection of top-$k1$ unique intents. It also builds dependencies of the frame length and output on the top-level intents. We find that these help the model to improve diversity and quality of top-\emph{k} output parses. To further validate the hypothesis, we devise two experiments to (1) identify the potential (i.e., upper limit of the accuracy) of the proposed NAR beam search and (2) examine the diversity and quality of the output parses compared to the baseline NAR beam search.

\subsection{Hybrid Teacher Forcing}

During our initial attempts to inspect the proposed model, we discovered that the inference performance is unstable. This was due to a discrepancy in the computation of the top-level intent during the training and inference. Recall that the outputs of the top-level intent module are logits (see Equation \ref{eq:proposed_nar_intent_conditioning}). In the inference, we use the logits to sample the top-$k1$ top-level intents, which are sparse one-hot vectors, and convert them back to the logits. Conversely, in the training, we do not sample top-level intents and instead use the direct outputs (dense logits) from the top-level intent module. We first attempted to resolve the problem with a teacher-forcing of the top-level intents. While this stabilized the inference, the training became unstable due to the sparsity of the teacher logits.

\begin{figure}[t]
    \centering
    \includegraphics[width=0.90\linewidth]{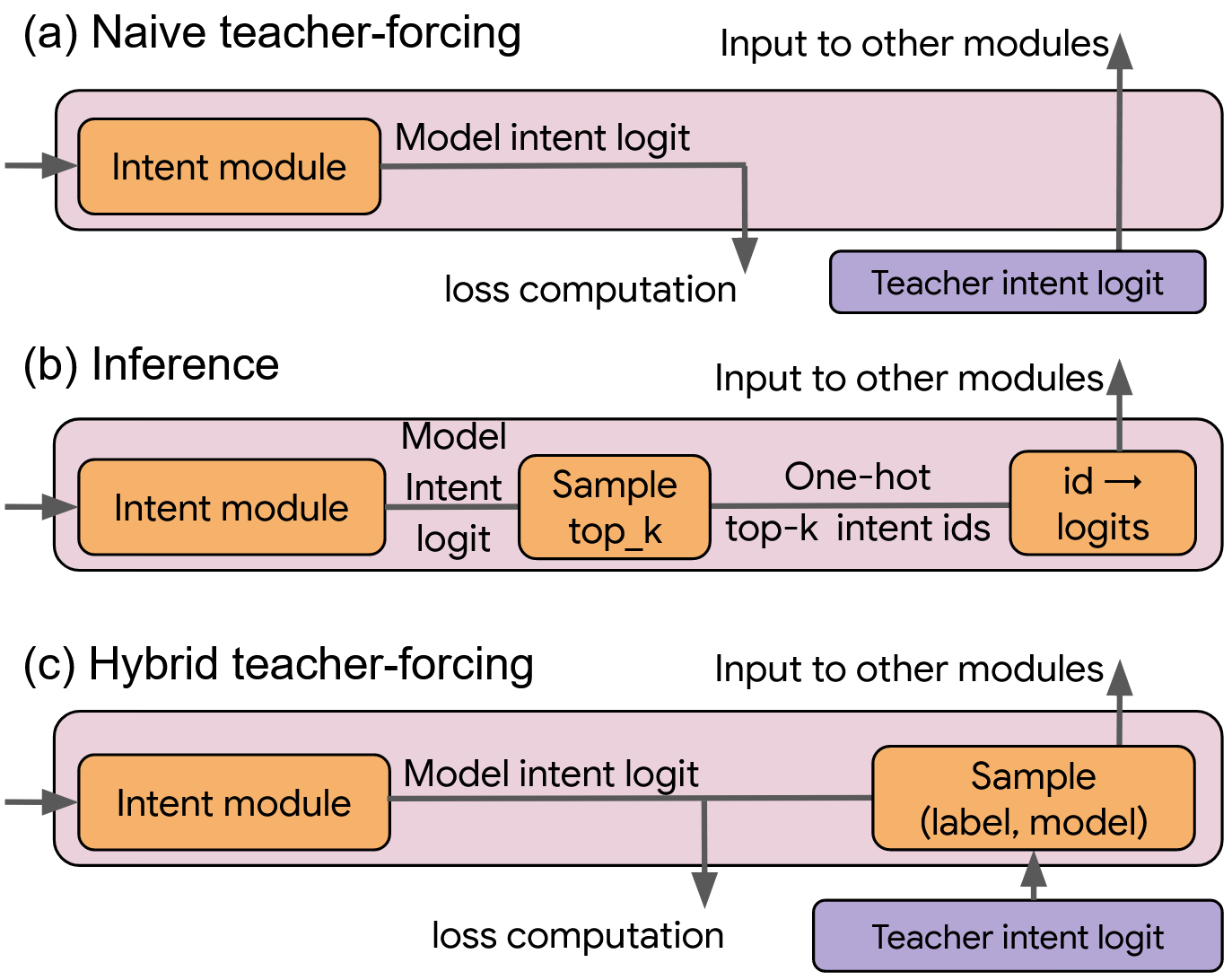}
    \caption{The computation of the top-level intent. (a) During training via naive teacher forcing, label intent logits are fed to the length module and decoder. (b) In the inference, top-\emph{k} intents are sampled from the model logits and fed to the other modules. (c) Hybrid teacher-forcing uses both label and model logits for training.}
    \label{fig:hybrid_teacher_forcing}
\end{figure}

We address this problem by leveraging a sampling strategy depicted in Figure \ref{fig:hybrid_teacher_forcing}. The idea is to uniformly sample logits from a pair of teacher and model logits as follows.

\vspace{-10pt}
\begin{equation}
\label{eq:hybrid_teacher_forcing}
\begin{aligned}
\mathrm{logits}(y_{1}) \sim \mathcal{U}(\mathrm{logits}(y_{1,\mathrm{label}}), \mathrm{logits}(y_{1,\mathrm{model}})).
\end{aligned}
\end{equation}

We name this strategy as \emph{hybrid teacher forcing} to reflect that the model leverages both teacher and model top-level intent logits. This sampling technique may be seen as a non-autoregressive variant of the scheduled sampling \citep{ref:scheduled_sampling}.

\subsection{Scoring Beam Search Outputs}
% 5. Method1 vs 2 vs 3(+length penalty)

A scoring method is used to select the top-\emph{k} parses from the $k1 \cdot k2$ pairs. We tested three methods. The first method $S_{1}$ uses the joint log probability of the top-level intent and frame length as the score.

\vspace{-10pt}
\begin{equation}
\label{eq:scoring_method1}
\begin{aligned}
S_{1} = \mathrm{log}(p(n,y_{1}|x)) = \mathrm{log}(p(n|y_{1},x)\cdot p(y_{1}|x)).
\end{aligned}
\end{equation}

The second method $S_{2}$ uses the joint log probability of the output, length, and top-level intent.

\vspace{-10pt}
\begin{equation}
\label{eq:scoring_method2}
\begin{aligned}
S_{2} = \mathrm{log}(p(y_{1:n},n|x)) = \\
\mathrm{log}\big(p(y_{2:n}|n,y_{1},x) \cdot p(n|y_{1},x) \cdot p(y_{1}|x)\big).
\end{aligned}
\end{equation}

The last scoring method $S_{3}$ combines the joint probability $p(y_{1:n},n|x)$ and a length penalty. 

\begin{equation}
\label{eq:scoring_method3}
\begin{aligned}
lp(y) = ((5+l)/6)^{\alpha}, \\
S_{3} = S_{2}(y)/lp(y),
\end{aligned}
\end{equation}

where $lp$ is the length penalty \citep{ref:google_mt_scoring}. 

Depending on $k1$, $k2$, and scoring method, the top-\emph{k} outputs may consist of parses with multiple distinct lengths and intents or parses with a single intent with multiple lengths.

\subsection{Semantic Parse Representation}
\begin{figure}[t]
    \centering
    \includegraphics[width=0.70\linewidth]{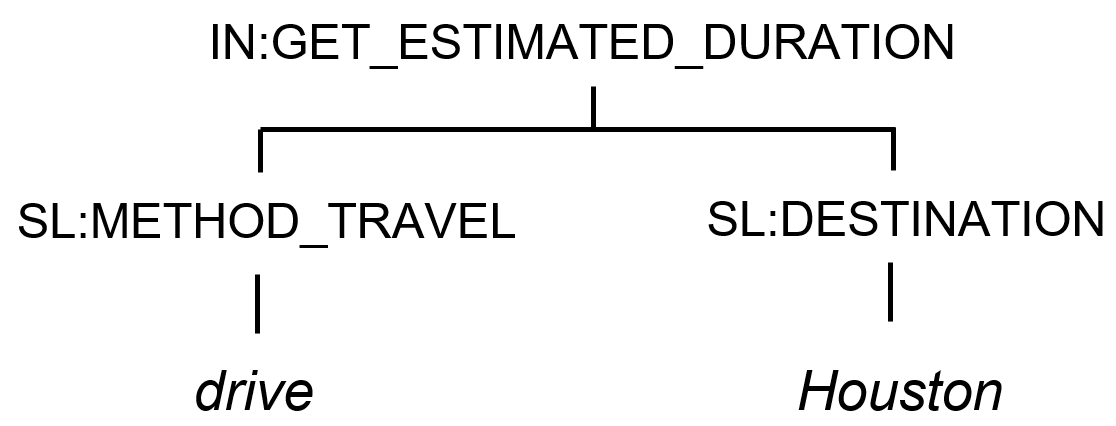}
    \caption{Decoupled representation of the semantic parse for a query ``How long is my drive to Houston?''.}
    \label{fig:decoupled_representation}
\end{figure}

%% TODO: pick a query and the corresponding parse for the figure below.
We represent semantic parses using the decoupled tree representation \citep{ref:SBTOP} that is an extension of the compositional tree representation \citep{ref:TOP}. An example is depicted in Figure \ref{fig:decoupled_representation}. Likewise, another query ``What is happening in Boston on New Year's Eve'' is semantically parsed into ``\textit{[in:get\_event [sl:location Boston ] [sl:date\_time on New Year's Eve ] ]}''.

An example of the decoupled tree representation is \emph{Index form} that replaces the copy tokens with the indices to the corresponding source tokens. In this sense, the above semantic parse is represented as ``\textit{[in:get\_event [sl:location 4 ] [sl:date\_time 5 6 7 8 ] ]}''. Together with the pointer network, this significantly reduces the size of output vocabulary \citep{ref:amazon_dont_parse_generate}. Another representation is \emph{Span form} introduced in \citep{ref:fb_span_pointer_network}. As opposed to specifying all indices of spans of copy tokens, the span form uses the start and end indices of the spans. That is, ``\textit{[in:get\_event [sl:location 4 4 ] [sl:date\_time 5 8 ] ]}''. The benefits of the span form include shorter target lengths and fixed number of leaf nodes that helps decoupling the syntax from the semantics. In addition, the number of valid frame length classes are halved as the frame lengths of parses in the span form are always even.

%%%%%%%%%%%%%%%%%%%%%%%%%%%%%%%%%%%%%%%%%%%%%%
\section{Experiments}
\label{sec:Experiments}
%%%%%%%%%%%%%%%%%%%%%%%%%%%%%%%%%%%%%%%%%%%%%%

\subsection{Datasets} 
% TOPv1, TOPv2
% + a table which shows statistics ?

To quantify the benefits of our NAR parsers over the baseline NAR, we utilize two conversational semantic parsing datasets. The first dataset is TOP (Task Oriented Parsing) \citep{ref:TOP}, a collection of human-generated queries in English and the corresponding semantic parses that are represented as hierarchical trees. Another public dataset we explore is TOPv2 \citep{ref:TOPv2}, which is an extension of TOP with six more domains.

\subsection{Evaluation Metrics}
% evaluation metrics are the following three.
% accuracy \& latency 

We mainly utilize \emph{exact match} (EM) to evaluate quality of both greedy decoding and beam decoding output. EM is defined as the percentage of queries whose label parses are correctly predicted.

\emph{Inference time} is the metric we use to quantify the latency benefit of our model. We measure the model inference time on 1 TPU \citep{ref:TPU} from the Google Cloud TPUv2 and 1 CPU of the Intel Cascade Lake CPU platform with 8GB RAM.
%and our TPU benchmark server uses an internal platform.
We used batch size 1 and computed latency on 1000 examples from the val set of the TOP dataset. We report per-example average latencies of these runs.

In addition, \emph{diversity} of the output parses is computed to quantify the capability of the proposed NAR approach at producing distinct top-\emph{k} outputs. We mainly compute two metrics: (1) number of unique top-level intents in top-\emph{k} output parses and (2) the distinct n-grams \citep{ref:distinct_n_gram}, which are commonly used diversity metrics in natural language processing \citep{ref:distinct_n_gram_ex1, ref:distinct_n_gram_ex2}. Specifically, we compute percentage of distinct n-grams by counting unique n-grams in top-\emph{k} outputs, dividing the counts by the total number of output tokens, and scaling it by 100.

\subsection{Implementation details}

\subsubsection{Encoder}
We use pre-trained $\mathrm{BERT_{BASE}}$ (L12/H768) as the encoders for all AR, baseline NAR, and proposed NAR for the experiments. We use uncased versions of them along with lowercased datasets as they generally performed better than the cased models with cased datasets. We leave it to future
work to quantify the performance with other pre-trained language models such as RoBERTa \citep{ref:RoBERTa}.

\subsubsection{Decoder}
We primarily compare the performances of two AR baseline decoders, the baseline NAR decoder, and the proposed NAR decoder. The two AR decoders are identical except that one is trained on the label parses represented in the index form and the other in the span form. All NAR decoders were trained on the label parses in the span form. We used the same decoder architecture for all models across all experiments. As discussed earlier, the baseline NAR is a replication of the NAR model proposed by \citet{ref:fb_span_pointer_network}. Specifically, we use the vanilla version that does not utilize the R3F loss \citep{ref:r3f}.

Unlike \citet{ref:fb_span_pointer_network} and \citet{ref:fb_NAR_lightconv_pointer} that use an MLP or CNN length module, our frame length module is a randomly initialized Transformer decoder; we found that it performs better. We used the same length module for both baseline and proposed NAR models. Similarly, we use a randomly initialized Transformer decoder for the top-level intent prediction. Further details of the model parameters are described in the appendix.

\subsubsection{Hyperparameters}
We observed that the ratio of the loss terms is important. In general, higher top-level intent penalty and moderate frame length penalty resulted in better accuracy. We use the hybrid teacher forcing to train the proposed NAR for the experiments presented in this paper. We trained the models using Adam optimizer \citep{ref:Adam} with exponential learning rate decay. Additionally, we use 1000 learning rate warmup steps. We also apply dropout to the source embeddings and the Transformer decoder for better generalization. The details of the parameters are described in the appendix.

%%%%%%%%%%%%%%%%%%%%%%%%%%%%%%%%%%%%%%%%%%%%%%
\section{Results}
\label{sec:results}
%%%%%%%%%%%%%%%%%%%%%%%%%%%%%%%%%%%%%%%%%%%%%%

We show that the proposed intent-conditioned NAR further closes the gap between AR and NAR SP. Most importantly, our model greatly improves the quality of top-\emph{k} output parses while maintaining the $O(1)$ decoding time of the baseline NAR.

We first present the greedy decoding results. We then investigate the upper-bound beam search performance of the baseline and proposed NAR. Next, we share the beam search results that quantify the quality and diversity of top-\emph{k} output parses. Lastly, we share the results of the comparison of various beam scoring methods. We point out that we used the identical encoder \& decoder architectures and model parameters for all AR \& NAR models. In addition, the identical frame length module is used for both baseline and proposed NAR models.

\begin{table}[t]
\begin{tabular}{
>{\arraybackslash}m{1.8cm}
>{\centering\arraybackslash}m{1.00cm} 
>{\centering\arraybackslash}m{1.00cm} 
>{\centering\arraybackslash}m{0.90cm}
>{\centering\arraybackslash}m{0.9cm} }
\toprule
Model & Latency, TPU & Latency, CPU & EM, TOPv1 & EM, TOPv2 \\
\toprule
Base. NAR & 1x & 10.26x & 82.56 & 84.86 \\

Prop. NAR & 1.01x & 11.17x & 83.11 & 85.22 \\
\midrule
AR, index  & 2.23x & 48.70x & 83.43 & 85.40 \\

AR, span   & 2.26x & 46.98x & 83.40 & 85.56 \\
\bottomrule
\end{tabular}
\caption{Greedy decoding results with BERT encoder. Latency numbers are measured with TOPv1 dataset and relative to the baseline NAR latency on TPU.}
\label{Table:greedy_decoding_results}
\end{table}

\subsection{Greedy Decoding}
% results 1: greedy decoding, 
% Methods: AR vs NAR+flen vs NAR+flen+1stintent
% Datasets: topv1, topv2
% Models: BERT-base, BERT-large, RoBERTa-base, RoBERTa-large
% with/without LS

Table \ref{Table:greedy_decoding_results} presents the greedy decoding results. At each decoding step of AR models, the most probable token is selected. This repeats until an END token is produced. In NAR greedy decoding, beam width of 1 is used for both frame length and top-level intent. No scoring method is used.

% \begin{table}[h]
% \caption{Greedy decoding results with $\mathrm{BERT_{LARGE}}$ encoder. Latency numbers were measured with TOPv1 dataset.}
% \label{Table:greedy_bert_large}
% \begin{tabular}{
% >{\arraybackslash}m{2.2cm}
% >{\centering\arraybackslash}m{1.1cm} 
% >{\centering\arraybackslash}m{1.1cm} 
% >{\centering\arraybackslash}m{0.95cm}
% >{\centering\arraybackslash}m{0.95cm} }
% \toprule
%  & Latency, TPU & Latency, CPU & EM, TOPv1 & EM, TOPv2 \\
% \toprule
% NAR, baseline  & 7.66ms & 138ms & ? & ? \\
% \midrule
% NAR, ours   & 9.19ms & 139ms & ? & ? \\
% \toprule
% AR, index form  & 15.6ms & 372ms & 83.34 & 85.78 \\
% \midrule
% AR, span form   & 15.6ms & 357ms & 83.83 & 85.94 \\
% \bottomrule
% \end{tabular}
% \end{table}

The results indicate that our model improves the performance of baseline NAR by 0.4-0.5 EM while matching the latency of the prior NAR approach. Compared to AR baselines, our model cuts the greedy decoding inference time (i.e., latency of the semantic parser including both encoder and decoder) by 4.2-4.4x on CPU and 2.2x on TPU.

\begin{table}[t]
\begin{tabular}{
>{\arraybackslash}m{5.2cm}
>{\centering\arraybackslash}m{0.1cm}
>{\centering\arraybackslash}m{1.4cm} }
\toprule
Models & \multicolumn{2}{c}{Oracle EM} \\
\toprule
\multicolumn{3}{c}{Variants of the baseline NAR} \\
\midrule
Shallow decoder & \multicolumn{2}{c}{83.34} \\

Shallow decoder, Label smoothing & \multicolumn{2}{c}{82.57} \\

Deep decoder & \multicolumn{2}{c}{83.73} \\

Deep decoder, Label smoothing & \multicolumn{2}{c}{83.03} \\
\midrule
Best of the baseline NAR & \multicolumn{2}{c}{\textbf{83.73}} \\
\toprule
\multicolumn{3}{c}{Variants of the proposed NAR}\\
\midrule
Shallow decoder & \multicolumn{2}{c}{86.22} \\

Shallow decoder, Label smoothing & \multicolumn{2}{c}{86.40} \\

Deep decoder & \multicolumn{2}{c}{86.89} \\

Deep decoder, Label smoothing & \multicolumn{2}{c}{85.66} \\
\midrule
Best of the proposed NAR &  \multicolumn{2}{c}{\textbf{86.89}} \\
\bottomrule
\end{tabular}
\caption{Empirical upper-bound of beam search with NARs. To obtain the oracle EMs, we feed the gold frame length for the baseline NARs in the inference. Likewise, we use the gold top-level intent for the proposed NARs.}
\label{Table:upper_bound_bert_base}
\end{table}

\subsection{Potential Impact of Intent Conditioning for Beam Search}
% \ag{was the k2=1 after using the gold intent for the proposed beam search ?}
% \GS{GS: yes, for beam search with the proposed model, I used k2=1 and oracle top-level intent}

% results 2: upper bound study. (motivates 1st-intent beam decoding)
% Methods: NAR+flen(when oracle flen is given) vs NAR+flen+1stintent(when oracle 1st intent is given)
% Datasets: topv1, topv2
% Models: BERT-base (shallow, no LS), BERT-base (shallow, LS), BERT-base (deep, no LS),  BERT-base (deep, LS), BERT-large (deep)

% \begin{table*}[t]
% \caption{Beam decoding results. The proposed NAR outperforms the baseline NAR in top-3 EM by 2.4 points. Compared to the AR models, our approach speeds up beam search inference by $6.7$ times on CPU.}
% \label{Table:beam_bert_base}
% \begin{tabular}{
% >{\arraybackslash}m{2.5cm}
% >{\centering\arraybackslash}m{2.0cm} 
% >{\centering\arraybackslash}m{2.0cm} 
% >{\centering\arraybackslash}m{1.6cm}
% >{\centering\arraybackslash}m{1.6cm}
% >{\centering\arraybackslash}m{1.6cm}
% >{\centering\arraybackslash}m{1.5cm}
% >{\centering\arraybackslash}m{1.5cm} }
% \toprule
%  & Latency, TPU & Latency, CPU & top-1 EM & top-2 EM & top-3 EM & top-1 IM & top-3 IM \\
% \toprule
% Baseline NAR  & 5.29ms & 63.4ms & 82.61 & 83.44 & 83.62 & 94.59 & 95.22\\

% Proposed NAR   & 5.91ms & 79.0ms & 83.12 & 85.15 & \textbf{86.00} & 94.85 & 98.05 \\
% \toprule
% AR, index form  & 14.7ms & 525ms & 83.44 & 86.46 & 87.34 & 94.61 & 96.44 \\

% AR, span form   & 15.0ms & 528ms & 83.47 & 86.22 & 87.12 & 94.67 & 96.22  \\
% \bottomrule
% \end{tabular}
% \end{table*}

\begin{table*}[t]
\begin{tabular}{
>{\arraybackslash}m{2.5cm}
>{\centering\arraybackslash}m{2.1cm}
>{\centering\arraybackslash}m{2.1cm}
>{\centering\arraybackslash}m{1.6cm}
>{\centering\arraybackslash}m{1.6cm}
>{\centering\arraybackslash}m{1.55cm}
>{\centering\arraybackslash}m{1.55cm} }
\toprule
 & TPU Latency & CPU Latency & top-1 EM & top-3 EM & top-1 IM & top-3 IM \\
\toprule
Baseline NAR  & 1x & 11.98x & 82.61 & 83.62 & 94.59 & 95.22\\

Proposed NAR   & 1.12x &  14.93x & 83.12 & 86.00 & 94.85 & 98.05 \\
\toprule
AR, index form  & 2.78x & 99.24x & 83.44 & 87.34 & 94.61 & 96.44 \\

AR, span form   & 2.84x & 99.81x & 83.47 & 87.12 & 94.67 & 96.22  \\
\bottomrule
\end{tabular}
\caption{Beam decoding results. The proposed NAR outperforms the baseline NAR in top-3 EM by 2.4 and top-3 IM by 2.8 points. Compared to AR models, our model cuts the beam search latency by $6.7$ times on CPU.}
\label{Table:beam_bert_base}
\end{table*}

% \begin{table*}[t]
% \caption{Beam decoding results. The proposed NAR outperforms the baseline NAR in top-3 EM by 2.4 and top-3 IM by 2.8 points. Compared to AR models, our approach cuts a beam search inference latency by $6.7$ times on CPU.}
% \label{Table:beam_bert_base}
% \begin{tabular}{
% >{\arraybackslash}m{2.4cm}
% >{\centering\arraybackslash}m{2.1cm} 
% >{\centering\arraybackslash}m{2.1cm} 
% >{\centering\arraybackslash}m{1.3cm}
% >{\centering\arraybackslash}m{1.3cm}
% >{\centering\arraybackslash}m{1.3cm}
% >{\centering\arraybackslash}m{1.25cm}
% >{\centering\arraybackslash}m{1.25cm} }
% \toprule
%  & TPU Latency & CPU Latency & top1 EM & top2 EM & top3 EM & top1 IM & top3 IM \\
% \toprule
% Baseline NAR  & 1x & 11.98x & 82.61 & 83.44 & 83.62 & 94.59 & 95.22\\

% Proposed NAR   & 1.12x &  14.93x & 83.12 & 85.15 & 86.00 & 94.85 & 98.05 \\
% \toprule
% AR, index form  & 2.78x & 99.24x & 83.44 & 86.46 & 87.34 & 94.61 & 96.44 \\

% AR, span form   & 2.84x & 99.81x & 83.47 & 86.22 & 87.12 & 94.67 & 96.22  \\
% \bottomrule
% \end{tabular}
% \end{table*}

We designed an experiment to investigate the potential benefits of the proposed beam search. We aimed to empirically quantify the upper bound of the beam search performance of the baseline NAR and our intent-conditioned NAR. In the experiment, we first trained the baseline and proposed NAR parsers, then fed the oracle (i.e., gold) frame length to the baseline NAR or the oracle top-level intent to the proposed NAR during the inference. To ensure validity of the results, we ran the experiments with different model configurations.

As shown in Table \ref{Table:upper_bound_bert_base}, the proposed NAR consistently scored higher oracle EM across various model configurations. It should be noted that the proposed NARs only use the gold top-level intent (i.e., the gold length is not used for the inference with the proposed NARs).
Table \ref{Table:beam_bert_base} confirms these results and shows that our NAR performs more effective beam-coding and achieves higher exact match for the top-\emph{k} output.

\subsection{Beam search}
% results 3: beam decoding, AR vs NAR+flen vs NAR+flen+1stintent
% Methods: AR vs NAR+flen vs NAR+flen+1stintent
% Datasets: topv1, topv2
% Models: BERT-base, BERT-large, RoBERTa-base, RoBERTa-large
% Top3 vs top2 vs top1
% setting: beam-flen 3, beam-1stintent 5. 

% beam search results with BERT_large is muted. (redundant)

% \begin{table*}[h]
% \caption{Beam decoding results with $\mathrm{BERT_{LARGE}}$ encoder on TOPv1 dataset.}
% \label{Table:beam_bert_large}
% \begin{tabular}{
% >{\arraybackslash}m{2.5cm}
% >{\centering\arraybackslash}m{2.0cm} 
% >{\centering\arraybackslash}m{2.0cm} 
% >{\centering\arraybackslash}m{1.6cm}
% >{\centering\arraybackslash}m{1.6cm}
% >{\centering\arraybackslash}m{1.6cm}
% >{\centering\arraybackslash}m{1.5cm}
% >{\centering\arraybackslash}m{1.5cm}}
% \toprule
%  & Latency, TPU & Latency, CPU & top-1 EM & top-2 EM & top-3 EM & top-1 IM & top-3 IM \\
% \toprule
% NAR, baseline  & 9.04ms & 146ms & 82.85 & 83.29 & 83.47 \\
% \midrule
% NAR, ours   & 8.57ms & 162ms & 82.90 & 84.68 & 85.22 \\
% \toprule
% AR, index form  & 17.2ms & 614ms & 83.16 & 86.50 & 87.52 \\
% \midrule
% AR, span form   & 17.9ms & 614ms & 83.80 & 86.73 & 87.82 \\
% \bottomrule
% \end{tabular}
% \end{table*}

In table \ref{Table:beam_bert_base}, we report the beam search results on TOP dataset as top-\emph{k} EM for $k={1,2,3}$. If the label parse matches with any of the top-\emph{k} output parses, it counts towards the top-\emph{k} EM. We also report top-level intent match (IM) percents. The top-\emph{k} parses are selected from the $k1 \cdot k2$ beam search outputs using the scoring method 3 (Equation \ref{eq:scoring_method3}). We used the same $k1 \cdot k2$ for the fair comparison. Specifically, we used $k2=25$ for the baseline NAR and $k1=25$, $k2=1$ for our NAR as there exist 25 distinct length classes and 25 distinct intent classes in TOP dataset. For the proposed NAR, we observed that a higher $k2$ has minor impact on top-\emph{k} EMs, compared to a higher $k1$. We note that the hybrid teacher forcing (Equation \ref{eq:hybrid_teacher_forcing}) was used for the training and helped the models achieve consistent test set accuracy.

The results demonstrate a large gap between AR and the existing NAR models in top-\emph{k} outputs. Notably, the top-3 EM from the existing NAR model is outperformed by the baseline AR by 3.5 points. Secondly, the proposed NAR outperforms baseline NAR in top-3 EM by 2.4 and top-3 IM by 2.8 points. Lastly, we show that the proposed NAR reduces the performance gap from the autoregressive models. The inference time of our NAR beam search is 6.7x times shorter on CPU and 2.5x times shorter on TPU. Our AR baselines manage parallel compute loads better than NAR models and thus result in smaller latency gaps on TPU.

\subsection{Diversity}

% results 4: diversity
% Methods: NAR+flen vs NAR+flen+1stintent
% Datasets: topv1, topv2
% Models: BERT-base
% comparison: average number of unique parses in top-3 output parses.

Table \ref{Table:diversity} elaborates on the diversity of the output parses with the baseline and proposed NAR models. Specifically, we compute their average numbers of unique top-level intents and distinct n-gram \citep{ref:distinct_n_gram} percentages in top-3 output parses. We find that top-\emph{k} parses of the proposed NAR outperforms the baseline NAR in all diversity metrics.

\begin{table}[h]
\begin{tabular}{
>{\arraybackslash}m{4.65cm}
>{\centering\arraybackslash}m{1.0cm}
>{\centering\arraybackslash}m{1.05cm}}
\toprule
Metric (higher is more diverse) & Baseline NAR & Proposed NAR \\
\toprule
top-3 Exact Match & 83.62 & 86.00 \\
\midrule
\# of unique 1st intents in top-3 & 1.12 & 3.00  \\
\midrule
Distinct 1-gram, sentence-wise & 69.84 & 79.92 \\
\midrule
Distinct 2-gram, sentence-wise & 85.93 & 94.12 \\
\midrule
Distinct 1-gram, corpus-wise & 25.77 & 40.79  \\
\midrule
Distinct 2-gram, corpus-wise & 40.36 & 53.77 \\
\bottomrule
\end{tabular}
\caption{We measure the diversities of the top-3 parses of the baseline and proposed NAR models reported in Table \ref{Table:beam_bert_base}. We report sentence-wise (i.e., distinct n-grams in each parse) and corpus-wise distinct n-grams (i.e., distinct n-grams in each collection of top-3 parses).}
\label{Table:diversity}
\end{table}

We observed that the baseline NAR beam search often duplicates the most probable parse with repeated tokens and/or invalid syntax. This suggests that the frame length conditioning alone is not effective at producing top-\emph{k} outputs (see Table \ref{Table:baseline_NAR_failure} and \ref{Table:proposed_NAR_example_output} for examples). Table \ref{Table:diversity} quantitatively confirms the observation with lower sentence-wise and corpus-wise distinct n-gram scores of the baseline NAR, compared to the proposed NAR.

%%%%%%%%%%%%%%%%%%%%%%%%%%%%%%%%%%%%%%%%%%%%%%
\section{Conclusion}
%%%%%%%%%%%%%%%%%%%%%%%%%%%%%%%%%%%%%%%%%%%%%%

% Summary of the findings \& contributions

We present a novel non-autoregressive neural semantic parser: the intent-conditioned NAR for improved top-\emph{k} decodings. The proposed model addresses the main limitation of the recent NAR models (i.e., the limited beam search capability) by decoupling the prediction of the top-level intent from output. This builds explicit dependencies of frame lengths and output parses on top-level intents and helps NAR semantic parsers to better control beam search. The proposed NAR model further closes the performance gap from AR models by improving the quality and diversity of top-\emph{k} outputs. We highlight that the \emph{decoding} time complexity is $O(1)$, regardless of output length or beam width.

% Future works include investigating generalization capability of the proposed approach.

% Consider this (maybe mention in the conclusion?): The dataset is not annotated with multiple outputs for ambiguous cases, so it does not really seem capable of measuring the ability of a semantic parser to output multiple parses in the cases of unresolvable ambiguity, aside from computing recall@k, which can be motivated differently.

% maybe mention the lack of datasets which can evaluate diversity & multi-modalities of semantic parsers.

% employing re-ranker can be mentioned as a future work.

\bibliography{anthology,custom}
\bibliographystyle{acl_natbib}

%%%%%%%%%%%%%%%%%%%%%%%%%%%%%%%%%%%%%%%%%%%%%%
\appendix
\clearpage

%%%%%%%%%%%%%%%%%%%%%%%%%%%%%%%%%%%%%%%%%%%%%%
\section{Comparisons with Other Benchmarks}
\label{sec:comparison_other_baselines}
%%%%%%%%%%%%%%%%%%%%%%%%%%%%%%%%%%%%%%%%%%%%%%

% GS: since our focus isn't SOTA model, we may remove the table below.
% or we send it to the appendix.
% results 5: comparison with other baselines

We compare the AR and NAR models we used with some of the benchmarks in the literature for the TOP \& TOPv2 datasets. While there are benchmarks with other pre-trained encoders, we report the prior models with pre-trained BERT encoders to be consistent with the AR and NAR models used in this work (Table \ref{Table:perf_comparison}).

\begin{table}[h]
\begin{tabular}{
>{\arraybackslash}m{5.05cm}
>{\centering\arraybackslash}m{0.65cm} 
>{\centering\arraybackslash}m{0.85cm} }
\toprule
Encoders & TOP & TOPv2\\ 
\toprule
\multicolumn{3}{c}{Non-autoregressive Models} \\
\toprule
$\mathrm{BERT_{BASE}}$ baseline (greedy) & 82.56 & 84.86 \\
\midrule
$\mathrm{BERT_{BASE}}$ \textbf{proposed} (greedy) & 83.11 & 85.22 \\
\toprule
% \multicolumn{4}{c}{Autoregressive Models} \\
\multicolumn{3}{c}{Autoregressive Models}   \\ 
\toprule
$\mathrm{BERT_{BASE}}$ (beam size = 4) \citep{ref:amazon_dont_parse_generate} & 83.13 & -  \\
\midrule
$\mathrm{BERT_{BASE}}$ (beam size = 5) \citep{ref:sp_compositional_code_embedding} & 85.01 & -  \\
\midrule
Transformer + $\mathrm{BERT_{BASE}}$ \citep{ref:rerank_bert_graph} & 82.51 & -  \\
\midrule
$\mathrm{BERT_{BASE}}$ (greedy-ours, index) & 83.43 & 85.40 \\
\midrule
$\mathrm{BERT_{BASE}}$ (greedy-ours, span)  & 83.40 & 85.56 \\
\bottomrule
\end{tabular}
\caption{Reported performance of semantic parsers with BERT encoders on TOP and TOPv2 datasets. Greedy refers to greedy decoding (i.e., beam size 1).}
\label{Table:perf_comparison}
\end{table}

We note that the two baseline AR semantic parsers used in this work are based on the AR architecture presented in \citet{ref:amazon_dont_parse_generate}. Our AR baselines perform comparably against the reported numbers in the literature, as shown in Table \ref{Table:perf_comparison}.

%%%%%%%%%%%%%%%%%%%%%%%%%%%%%%%%%%%%%%%%%%%%%%
\section{Scoring Beam Outputs}
\label{sec:results_beam_scoring}
%%%%%%%%%%%%%%%%%%%%%%%%%%%%%%%%%%%%%%%%%%%%%%

% results 5: scoring methods
% Methods: NAR+flen+1stintent
% Datasets: topv1,
% Models: BERT-base
% Top3, top2, top1
% setting: beam-flen 3, beam-1stintent 5.

Table \ref{Table:scoring} depicts the performance of the three scoring methods on the TOP dataset. Each scoring method sorts the beam outputs of the proposed NAR differently (Equation \ref{eq:scoring_method1}-\ref{eq:scoring_method3}).

\begin{table}[h]
\begin{tabular}{
>{\arraybackslash}m{2.6cm}
>{\centering\arraybackslash}m{1.1cm}
>{\centering\arraybackslash}m{1.1cm}
>{\centering\arraybackslash}m{1.1cm}}
\toprule
\multirow{2}{*}{Scoring Method} & \multicolumn{3}{c}{Exact Match} \\
\cmidrule{2-4}
 & top-1 & top-2 & top-3 \\
\toprule
$S_{1}$ (Equation \ref{eq:scoring_method1}) & 83.11 & 84.88 & 85.36 \\
\midrule
$S_{2}$ (Equation \ref{eq:scoring_method2}) & 83.11 & 84.95 & 85.49 \\
\midrule
\multicolumn{4}{l}{$S_{3}$ (Equation \ref{eq:scoring_method3})} \\
$\alpha$ = 1.0 & 83.13 & 85.21 & 85.81 \\
$\alpha$ = 3.0 & 83.12 & 85.15 & \textbf{86.00} \\
$\alpha$ = 5.0 & 83.15 & 85.08 & 85.93 \\
\bottomrule
\end{tabular}
\caption{Performance of the three scoring methods.}
\label{Table:scoring}
\end{table}

Our study indicates that the scoring method 3 works the best for our NAR model, empirically demonstrating the effectiveness of the length penalty \citep{ref:google_mt_scoring} for NAR models. We observed that high length penalties (e.g., $\alpha = 2.0 - 3.0$) often corresponded to the highest EM. For fair comparison, we also applied the length penalty with $\alpha = 1.0$ for the AR models for all experiments reported in this work.

%%%%%%%%%%%%%%%%%%%%%%%%%%%%%%%%%%%%%%%%%%%%%%
\section{Model Architecture Details}
\label{sec:Model_details}
%%%%%%%%%%%%%%%%%%%%%%%%%%%%%%%%%%%%%%%%%%%%%%

\begin{figure*}[h]
    \centering
    \includegraphics[width=\linewidth]{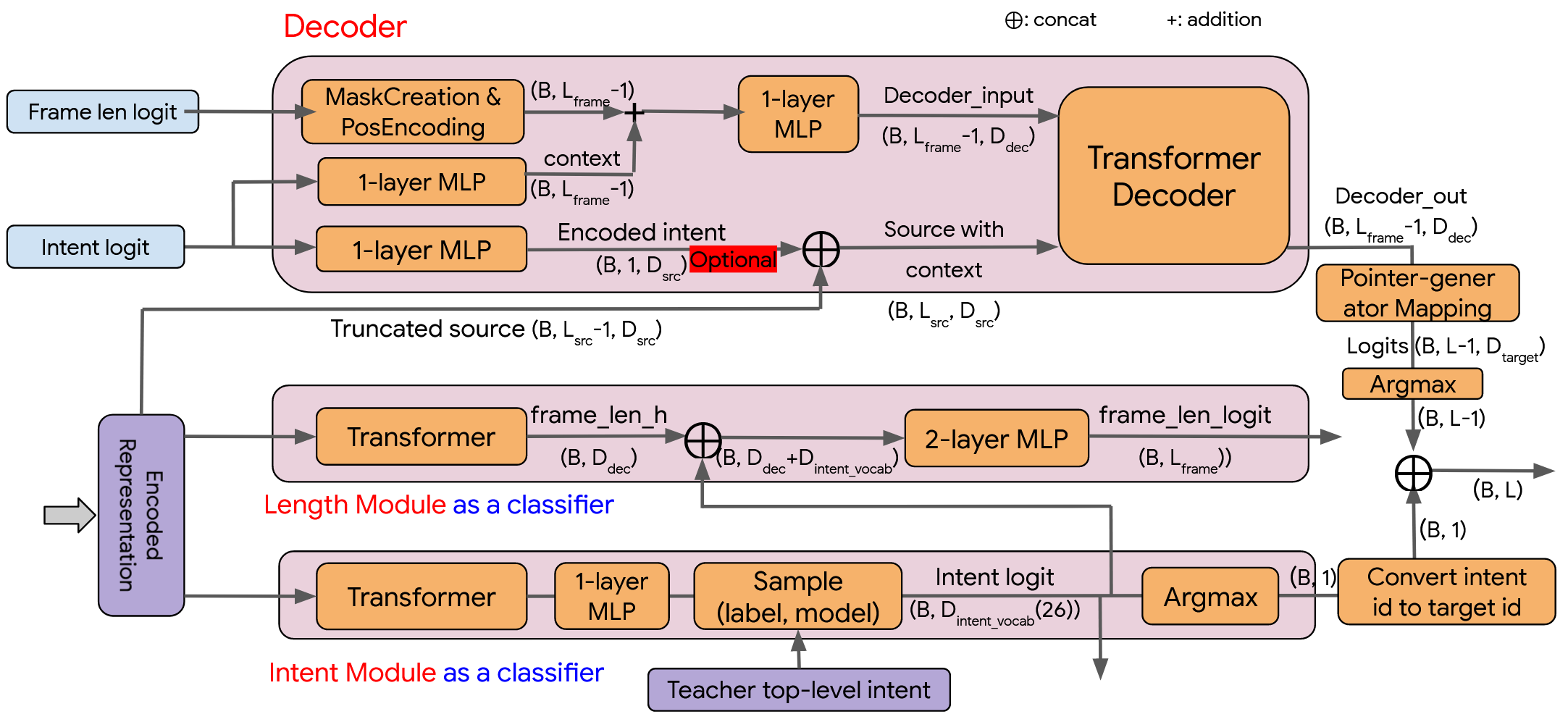}
    \caption{The detailed architecture of the proposed NAR.}
    \label{fig:proposed_model_details}
\end{figure*}

Here, we elaborate on the details of the model architecture and parameters used across different experiments. Table \ref{Table:model_parameters} specifies hyper-parameters we used to build the AR and NAR models used across different experiments. Compared to the AR models, the baseline NAR has roughly 12.5\% more model parameters as it includes the frame length module in addition. The proposed NAR has approximately 4.4\% more model parameters than the baseline NAR due to the addition of the intent module. Figure \ref{fig:proposed_model_details} depicts the detailed model architecture for the proposed NAR. The figure includes (sub)-modules, inputs, outputs, as well as the dimensionality of them.

\begin{table*}[ht]
\begin{tabular}{
>{\centering\arraybackslash}m{5.5cm}
>{\centering\arraybackslash}m{2.9cm}
>{\centering\arraybackslash}m{2.9cm} 
>{\centering\arraybackslash}m{2.9cm}}
\toprule
 & AR & baseline NAR & proposed NAR \\
\toprule
Number of parameters & 114.1M & 128.4M & 134.1M \\
Number of TPUs used for training & \multicolumn{3}{c}{8 Google Cloud TPUv2} \\
decoder (layer/hidden/head) & \multicolumn{3}{c}{L4/H256/HD2} \\
length module (layer/hidden/head) & N/A & \multicolumn{2}{c}{L8/H256/HD4} \\
intent module (layer/hidden/head) & N/A & N/A & L8/H256/HD4 \\
nonlinearity & \multicolumn{3}{c}{Relu} \\
model dropout & \multicolumn{3}{c}{0.0316} \\
source embeddings dropout & \multicolumn{3}{c}{0.0022} \\
\midrule
$\mathrm{\lambda_{length}}$  & N/A & \multicolumn{2}{c}{10} \\
$\mathrm{\lambda_{intent}}$  & N/A & N/A & 100 \\
\midrule
optimizer & \multicolumn{3}{c}{Adam} \\
learning rate & \multicolumn{3}{c}{0.00004} \\
learning rate warmup steps & \multicolumn{3}{c}{1000} \\
learning rate scheduler & \multicolumn{3}{c}{Exponential Decay} \\
batch size & \multicolumn{3}{c}{256} \\
\bottomrule
\end{tabular}
\caption{Hyper-parameters related to the models used across different experiments presented in this work.}
\label{Table:model_parameters}
\end{table*}

% \begin{figure}[h]
%     \centering
%     \includegraphics[width=\linewidth]{Figures/proposed_model.PNG}
%     \caption{(maybe redundant, may be used later) Intent-conditioned NAR model diagram}
%     \label{fig:proposed_model}
% \end{figure}

\end{document}